# A Hybrid Model for Enhancing Lexical Statistical Machine Translation (SMT)

**Ahmed G. M. ElSayed [1], Ahmed S. Salama [2] and Alaa El-Din M. El-Ghazali [3]**

**[1]Computer and Information Systems Department, Sadat Academy for Management Sciences (SAMS)
Cairo, 002, Egypt**

**[2] Computer and Information Systems Department, Sadat Academy for Management Sciences (SAMS)
Cairo, 002, Egypt**

**[3] Computer and Information Systems Department, Sadat Academy for Management Sciences (SAMS)
Cairo, 002, Egypt**

**Abstract**

The interest in statistical machine translation systems increases currently due to political and social events in the world. A proposed Statistical Machine Translation (SMT) based model that can be used to translate a sentence from the source Language (English) to the target language (Arabic) automatically through efficiently incorporating different statistical and Natural Language Processing (NLP) models such as language model, alignment model, phrase based model, reordering model, and translation model. These models are combined to enhance the performance of statistical machine translation (SMT). Many implementation tools have been used in this work such as Moses, Gizaa++, IRSTLM, KenLM, and BLEU. Based on the implementation, evaluation of this model, and comparing the generated translation with other implemented machine translation systems like Google Translate, it was proved that this proposed model has enhanced the results of the statistical machine translation, and forms a reliable and efficient model in this field of research.

## 1. Introduction

Statistical Machine Translation (SMT) deals with automatically mapping sentences in one human language into another human language. Therefore, it translates from the source language to the target language. The goal of SMT is to analyze automatically existing human sentence translations, to build general translation rules. The problem of machine translation has not been solved yet. Much research and development still needed to earn the reliability of humans by preforming a fluency translation as humans do. Therefore, a method represented to tune the translation parameters and improve the translation quality.

## 2. Literature Review

In general, many models have been used in machine translation for many years. SMT was the preferred approach for research and development of machine translation systems within this period. Rabeh Zbib conducted a research in 2010 for how to use linguistic knowledge in statistical machine translation in MIT, USA [1]. In 2007, Sarikaya and Deng published about joint morphological-lexical language modeling for machine translation in the conference of the North American Chapter of the Association for Computational Linguistics about [2]. Rabih Zbib, Spyros Matsoukas, Richard Schwartz, and John Makhoul.in 2010 conducted a research for using decision trees for lexical smoothing in statistical machine translation [3]. Ahmed Ragab Nabhan and Ahmed Rafea in 2005 have done a research on tuning statistical machine translation parameters in central laboratory for agricultural expert system (CLAES) [4]. Philipp Koehn in 2004 presented pharaoh: a beam search decoder for phrase-based statistical machine translation models [5]. Papineni, K.; Roukos, S.; Ward, T.; Zhu, W. J. in 2002 introduced the "BLEU: which is a method for automatic evaluation of machine translation" in ACL 40th annual meeting of the association for computational linguistics [6]. Nizar Habash and Owen Rambow, in 2005 conducted a research about Arabic tokenization using part-of-speech tagging and morphological disambiguation in one fell swoop [7]. Ibrahim Badr, Rabih Zbib, and James Glass in 2008 done a research on segmentation for English-to-Arabic statistical machine translation [8]. Franz Och and Hermann Ney have done a research in 2000 about the improved statistical alignment models in ACL [9]. Ibrahim Badr, Rabih Zbib, and James Glass in 2009 have conducted a research for syntactic phrase reordering for English-to-Arabic statistical machine translation in MIT,

USA [10]. Kenneth Heafiel, Philipp Koehn, and Alon Lavie did a research in 2013 about grouping language model boundary words to speed K-Best extraction from hypergraphs in proceedings of the conference of the north American chapter of the association for computational linguistics (NAACL) [11]. In 2004 Noah A. Smith has done a research on Log-Linear Models in Johns Hopkins University in USA [12].

## 3. The Proposed Model Architecture

The proposed model for enhancing lexical statistical machine translation architecture is shown in Fig. 1 Model components are discussed briefly in the following subsection.

### 3.1 System Administrator

System Admin is the first component in this model, which will use a user interface and the DBMS tool to enter vocabularies, patterns, and the already translated sentences into the database to use it as a reference for translations.

### 3.2 User

The user is the person who will use the model to enter the sentence required for translation, and after he enters it, he will get the result and the translation for this sentence which has the best probability score.

### 3.3 User Interface (UI)

This is the user interface, which used from system admin or the user of this model. System admin will have interface, which allows him to control the model and administrate it while the user will have interface, which allows him to enter the required sentence for translation then get the translation for this sentence. Both of the system admin and the user will use the user interface to interact with the DB across the Database Management System.

### 3.4 The Database (DB)

This database will be used to store all data, which will be used in this model, plus the new translated sentences with their sequence. This component will contain the source English sentence before Arabic translation plus the Parallel corpus or corpora, which is the already translated sentences and patterns from human translations. It will also contain a dictionary or vocabularies and patterns for many and many words and sentences to get all possible translations for each word. The more human-translated patterns in the database, which the model can analyze, and use, the better the translation quality will be.

### 3.5 Database Management System (DBMS)

DBMS is the software, which allows system admin and users of this model to interact with this database. System admin will be allowed to enter new patterns or new sentences, and enter new translated words. User will be allowed to enter the sentence required for translation.

### 3.6 Data Preparator

This component focuses on converting the input parallel corpus into format, which is suitable for using in this model. Data preparator of this model performs four main functions as follows:

- Normalization by converting all words of source and target language to upper or lower cased in all sentences.
- Tokenization for all data in the both language corpus data by inserting putting spaces between words and punctuation.
- True-casing by generating probabilities for all words in the parallel corpus and building a model for this true-casing to be used for generating the file of the true-casing for each language.
- Cleaning data by delete long sentences, which are longer than specific number or remove empty sentence or misaligned sentences, which can affect the translation quality.

### 3.7 Trainer

This is the core of this model. In this component, many models combined to perform a hybrid model for data training to get better translation result. The trainer applies training on the following components:

#### 3.7.1 Language Model (LM)

Language model built to make sure that translation has been performed correctly with fluency, so it has been built in the target language. Language model built based on a combination of ngram models.
The “language model” or “lm” is a statistical description of one language that includes the frequencies of token-based n-grams occurrences in a corpus. Language model is trained from a large monolingual corpus and saved as a file. The language model file is a required component of every translation model. This model depends on a combination from the unigram language model to a 5gram language model and the following is an example of how the Bigram and Trigram language model was built [13].

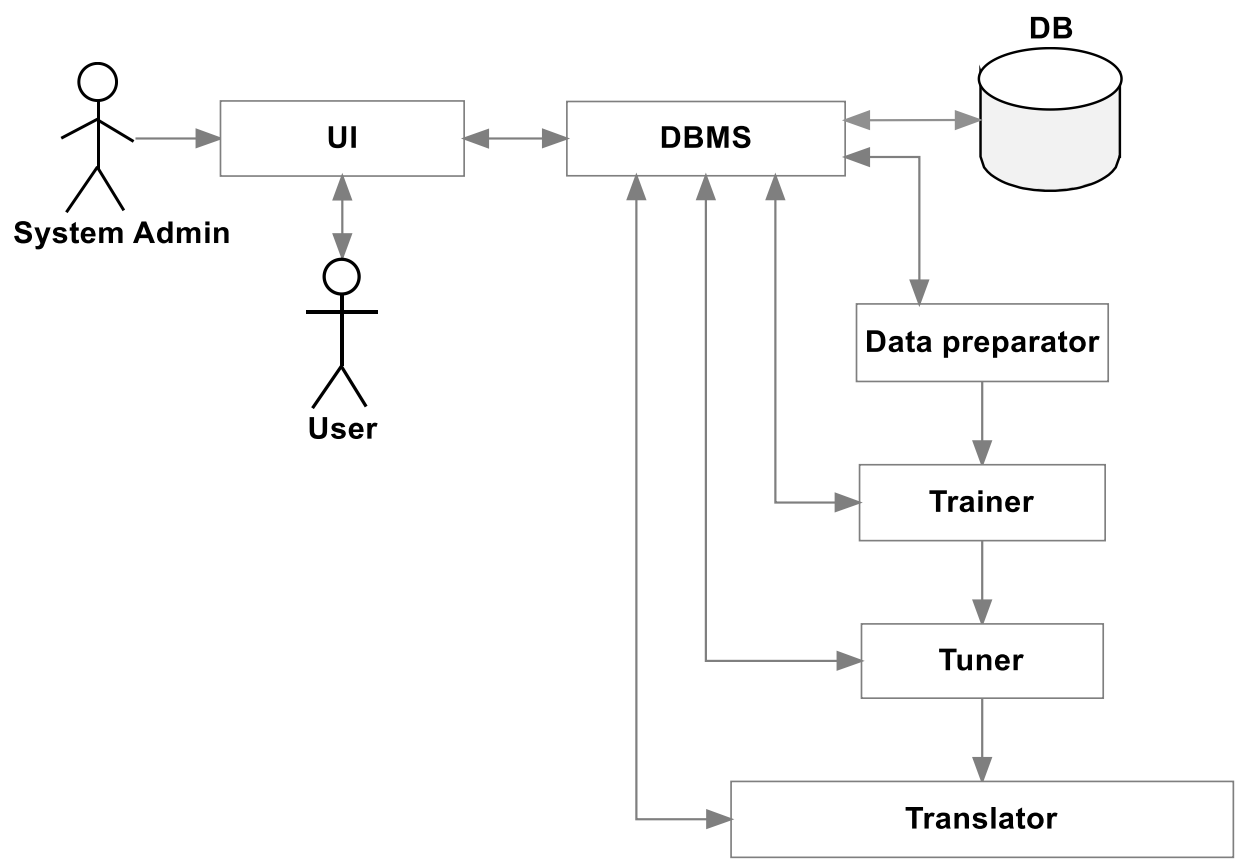

Fig. 1 Architecture Diagram of the hybrid Model for enhancing SMT.

Consider V is a finite set of all words in the language. Assume that $x_1, x_2, \ldots, x_n$ is the sequence of words in the language while $x_n > 1$ and $x_n$ is a special symbol STOP, and (STOP is not a member in the set V). Define V+ as a set of all sentences with the vocabulary V, this means that V+ is infinite set because sentences can be in any length. Consider a sequence of random variables $X_1;X_2, \ldots, X_n$. Each random variable can take any value in a finite set V, and n is the number of words + 1 (STOP symbol). A Markov bigram, and trigram language models are presented as follows:

- A Bigram Markov language model:

$P(X_1 = x_1;X_2 = x_2, \ldots, X_n = x_n)$

$= P(X_1 = x_1) \prod_{i=2}^{n} P(X_i = x_i | X_1 = x_1, \ldots \ldots, X_{i-1} = x_{i-1})$

$= P(X_1 = x_1) \prod_{i=2}^{n} P(X_i = x_i | X_{i-1} = x_{i-1})$ (1)

where $P(X_1 = x_1)$ is the probability of the first random variable $X_1$ to be equal to the first word $x_1$ in the target language, and so on. The first-order Markov assumption: for any $i \in \{2 \ldots n\}$, for any $x_1 \ldots x_i$,

$P(X_i = x_i | X_1 = x_1, \ldots \ldots, X_{i-1} = x_{i-1}) =$
$P(X_i = x_i | X_{i-1} = x_{i-1})$

- A Trigram Markov language model:

The Trigram Language Model consists of:

- A finite set $V$
- A parameter $q(w|u, v)$ for each trigram $u, v, w$ such that $w \in V \cup \{STOP\}$, and $u, v \in V \cup \{*\}$. Where the value for $q(w|u, v)$ is the probability of seeing the word $w$ after the bigram $u, v$. Therefore $q(w|u, v)$ defines a distribution over possible words $w$ conditioned on the bigram context $u, v$.

For any sentence $x_1 \ldots x_n$ where $x_i \in V$ for i=1…(n-1), and $x_n$=STOP, the probability of the sentence under the trigram language model is as follows:

$P(X_1 = x_1; X_2 = x_2; X_3 = x_3, \ldots, X_n = x_n)$

$= P(X_1 = x_1) \times P(X_2 = x_2 | X_1 = x_1) \times \prod_{i=3}^{n} P(X_i = x_i | X_{i-2} = x_{i-2}, X_{i-1} = x_{i-1})$

$= \prod_{i=1}^{n} P(X_i = x_i | X_{i-2} = x_{i-2}, X_{i-1} = x_{i-1})$ (2)

(For convenience we assume $x_0 = x_{i-1} = *$, where * is a special "start" symbol)

From **Error! Reference source not found.**), Eq. (1) is as follows:

$P(x_1 \ldots x_n) = \prod_{i=1}^{n} q(x_i | x_{i-2}, x_{i-1})$ (3)

Where we define x0 = xi-1 = *

For example:

يلعب أحمد الكرة STOP

We would have: $P(\text{يلعب أحمد الكرة}\ STOP)$

$= q(\text{يلعب} | *, *) \times q(\text{أحمد} | *, \text{يلعب}) \times q(\text{الكرة} | \text{يلعب}, \text{أحمد})$
$\times q(STOP | \text{أحمد}, \text{الكرة})$

A natural estimate (the "maximum likelihood estimate"):

$q(w_i | w_{i-2}, w_{i-1}) = \frac{Count(w_{i-2}, w_{i-1}, w_i)}{Count(w_{i-2}, w_{i-1})}$ (4)

### 3.7.2 Alignment Model

Alignment models used in statistical machine translation to determine relation between translation words in a sentence in one language compared to the words in a sentence with the same meaning in a different language. The alignment model form an important part of the translation process, as it's used to produce word-aligned text, which is used to create machine translation systems. IBM five models are famous models for alignment. They were proposed about 25 years ago from now but they still form the state of the art models for alignment. They consist of five models for word-to-word alignment called "IBM 1-5 models". With alignment model, a generalization of the procedures of extracting the phrase of phrase-based systems with its corresponding sequence can be done. Alignment models will be used to align words to enhance translations. IBM model 4 has been used because it has increased sophistication of model 3, so it sums over high probability alignments, and it search for the most probable alignment [13]. Aligned data are elements of a parallel corpus in two languages. Each element in one language matches the corresponding element in the other language. This model consists of the training model for inspecting the alignments for data and create alignment table which used later to align words and help in perform reordering model.

### 3.7.3 Translation and Phrases Extractor

This component used to get the direct translation for words then create table with maximum likelihood for words translations and extract phrases in one file with their scores. It extracts the phrase table, which is a statistical description of a parallel corpus of source-target language sentence pairs.

### 3.7.4 Reordering Model

After phrases extraction with their scores, the reordering model uses the alignment model to generate reordering table. Reordering table contains statistical frequencies that describe changes in word order between source and target language.

### 3.7.5 Translation Model

This is the final components generated as output after training all the previous components, which will be tuned later or can be used to generate the translation directly. The translation model contains translation table, alignment model, phrase table, reordering table, and language model, which will be used for translation.

### 3.8 Tuner

The main purpose of tuning the trained SMT model is to enhance statistical machine translation results and improve its quality. Since SMT, training uses a linear model, the tuning aims to find the optimal weights for this linear model and minimize error rating, where optimal weights are those, which maximize translation performance. Therefore, that tuning is a process of finding the optimized settings for the translation model. The tuning process translates thousands of source language phrases in the tuning set with a translation model, compares the model's output to a set of reference human translations, then it adjusts the settings with the intention to improve the translation quality. This process continues through iterations. With each iteration, the tuning process repeats the steps until it reaches an optimized translation quality and minimized error rate.

### 3.9 Translator

This is the final component of this model, which applies the translation model with all its components (language model, translation table, alignment model, phrase table, and reordering table) on sentence provided by user to translate it to the appropriate translation. The translator job is to implement translation model to the required sentence for translation and translate the sentence then send the translation back to user through the UI.

## 4. The proposed algorithms and activity diagrams for the proposed model

In the coming two subsections, activity diagrams and algorithms are proposed.

### 4.1 Proposed Model Activity Diagrams

In this section, two activity diagrams for this proposed model are shown in **Error! Reference source not found.** and **Error! Reference source not found.**. **Error! Reference source not found.** shows the activity diagram for the training and learning activities of the main model components. **Error! Reference source not found.** shows the steps starting from when user provides the source English sentence until the user receives the translated sentence in the target language (Arabic language).

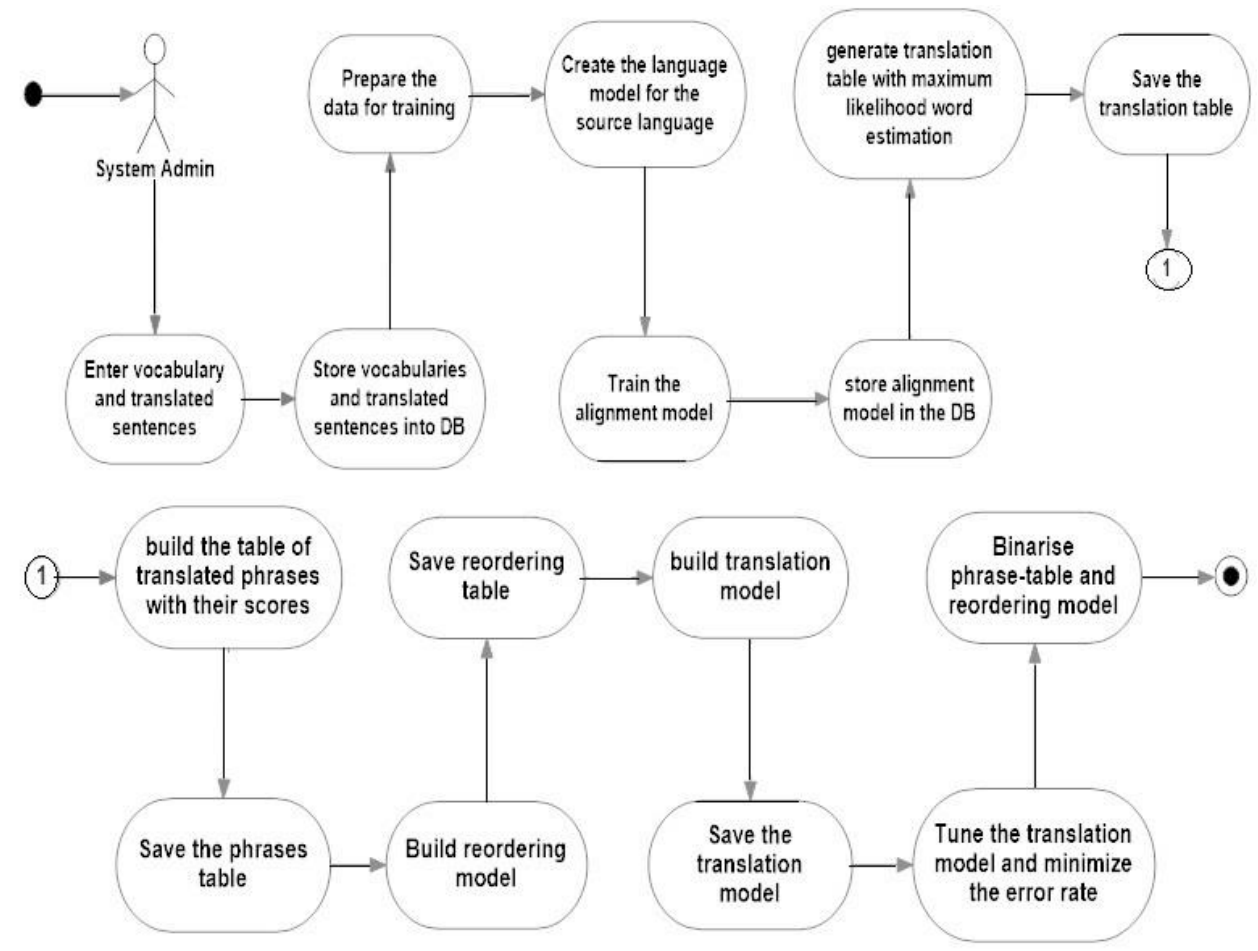

Fig. 2 Activity Diagram for Learning and Training the Model.

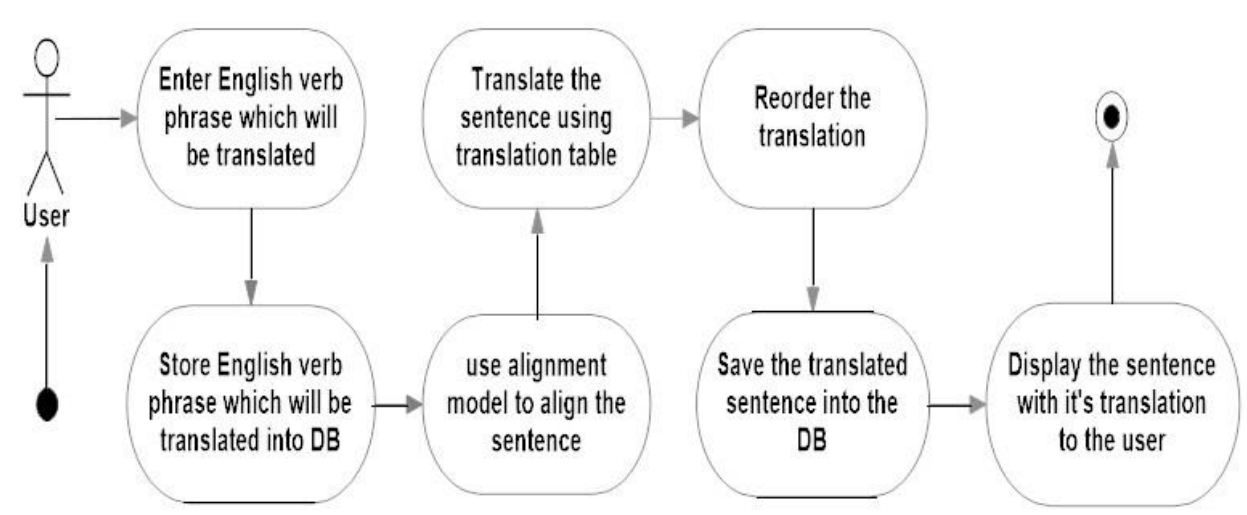

Fig. 2 Activity Diagram for the process of the translation using the model.

### 4.2 Model Algorithms

The proposed model algorithms for the model learning, training and the model translation are presented in this section.

#### 4.2.1 Model training and learning algorithm

1. System administrator will support the dictionary and feed it with vocabularies and translated sentences or parallel corpora, which used as a reference for the translation model. He will enter this data through the user interface (UI) and Database Management System (DBMS) tool.
2. Prepare the parallel corpora for the training process by normalizing words, tokenization, truecasing, and cleaning.
3. Create the language model for the source language to help in translation process.
4. Train the word alignment model with parallel corpus data.
5. Save the alignment model files for references and to help create the reordering model.
6. Generate translation table with maximum likelihood estimation.
7. Save the translation table.
8. Build phrases table for translated phrases with their scores.
9. Save the phrases table.
10. Build reordering model.
11. Save the reordering table.
12. Build the translation model.
13. Save the translation model.
14. Tune the translation model and minimize the error rate for translation to improve translation quality and add model upgrade.
15. Binarise the phrase table and reordering table.

By performing all the previous steps, the translation model would be trained for parallel corpus.

#### 4.2.2 The user source statement translation algorithm

1. User enters the English verb phrase statement, which will be translated through the UI and DBMS tool.
2. Store the English verb phrase to the DB.
3. Align the sentence using alignment model.
4. Translate the sentence using the translation table.
5. Reorder the translated sentence.
6. Save the translated sentence into the DB.
7. Display the translated sentence to the user.

After performing the previous steps, the required sentence for translation will be translated into the target language "the Arabic language" and displayed to the user of this model.

## 5. PROPOSED MODEL IMPLEMENTATION

This model presents integration between many components for enhancing lexical statistical machine translation including language model, alignment model, phrase based translation model, reordering model, translation modeling, and tuning the final translation model. All these components combined together in order to improve the quality of the output of statistical machine translation and build a more reliable model. Set of implementation tools were used to build and train the main components of the model. Data sets from different data sources were used during the training and learning phases of the model. The Implemented model was applied on the translation of test data set. The generated translation quality was measured by the BLEU score tool.

### 5.1 The Proposed Model Implementation Tools

This model used a various set of tools to generate the final model. These tools are as follows:

- IRSTLM Toolkit for language modeling.
- KenLM also for language modeling.
- In language modeling, both IRSTLM Toolkit and KenLM tool were used together combined with the main tool for language modeling and training. Both of them do the same purpose but the only difference is IRSTLM is better in query and editing the model, but KenLM is faster in model creation because it is multithreaded and doing the language model binarisation.
- GIZA++ word aligning tool to align the parallel corpus and train the alignment model.
- This toolkit is an implementation of the IBM models that started statistical machine translation research and the state of the art techniques.
- MOSES as a complete statistical machine translation system. In the rest of this model for training and tuning, Moses have been used which allows training translation models for language pairs. Once creating the translation model, an efficient search algorithm finds quickly the highest probability translation among the exponential number of choices.
- Bi-Lingual Evaluation Understudy (BLEU) score tool, which is the most famous scoring, and testing tool for statistical machine translation quality was used for model evaluation. The BLEU score indicates how closely the token sequences in one set of data for example in machine translation output, correlate with or match the token sequences in another set of data, such as a reference human translation.

### 5.2 Model training and learning

In the training and learning of the proposed model, different data sets were used. After the collection of these data sets, data preparation activities should be applied in these data sets before the model training, tuning, and evaluating. These preparation activities were presented in section 3.63.6. As input for this experiment to test this model, a small amount of data used from a collection of translated documents from the United Nation for model training. Model training data presented as a set of 74067 sentences (3138881 token) before data cleaning and 67575 sentences after data cleaning [14]. For tuning a collection of translated sentences from Tatoeba were used consists of 13000 sentence (65300 token) [15]. Finally for model evaluation a parallel corpus of Ubuntu localization files consist of 6000 sentences (24400 token) were used to get the BLEU result and assess the model [15].

### 5.3 Model Evaluation Result

After using the BLEU score tool in evaluating the implemented model, it was found that the score for this model before tuning was 8.99. This considered a good result compared to the number of data used for training in this model. In addition, the data used for the model evaluation is not from the same domain, which decreases the BLEU result. After tuning this model, the BLEU score shows a significant improvement as it achieved a result of 19.52. As known that the more data used in training the model, the better result we get. In case of using data from the same domain for testing, less errors would be achieved from the model and better translation BLEU result accomplished. When using small number of corpus, then the expected BLEU score is very low. As a conclusion for this model, it shows that after tuning the model, a better result achieved. The outcome for this model has jumped from 8.99 to 19.52, which is a significant improvement and a good result came from small sized corpus used in training. In addition, it's considered a good result since the test data set were collected from different domains rather than the same domain for training. As the model training is continued using more new data sets in the same domains, the better the translation quality it generates.

### 5.4 Model Translation Results Using Case Study

In the following sub sections, a set of English sentences were used as a case study to present the generated model Arabic translation before tuning and after tuning.

#### 5.4.1 English Sentences used for evaluation

- Britain has largely excluded itself from any leadership role in Europe.
- Many members of congress represent districts.
- I will transfer the money.
- Italy is far from Brazil.

#### 5.4.2 Model Translation before Tuning

- The Arabic translation for the sentence "Britain has largely excluded itself from any leadership role in Europe" before tuning as shown in Fig. 3 is "حد ذاتها لبريطانيا قد مستبعدة من أي الدور القيادى في أوربا"

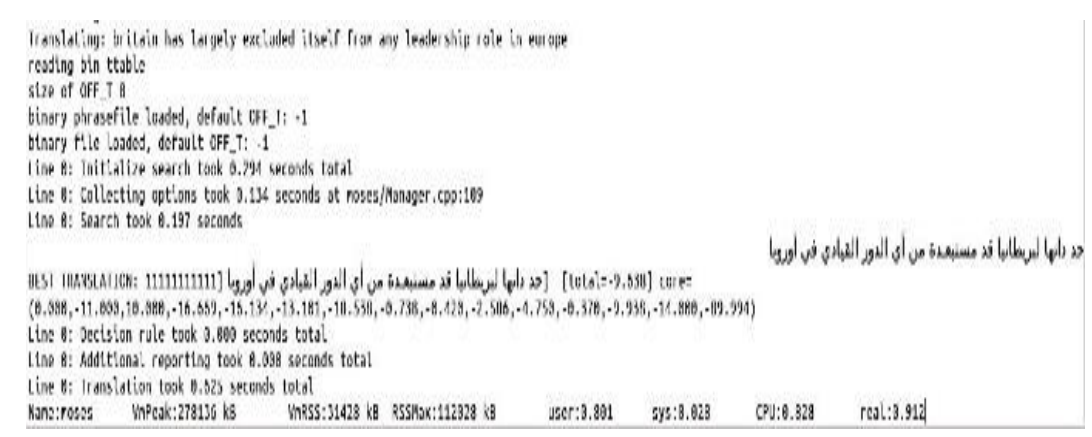

Fig. 3 Arabic translation of "Britain has largely excluded itself from any leadership role in Europe" before tuning.

- The Arabic translation for the sentence "Many members of congress represent districts" before tuning as shown in Fig. 4 is "العديد من الأعضاء عشر تمثل المستقرة والأمنة"

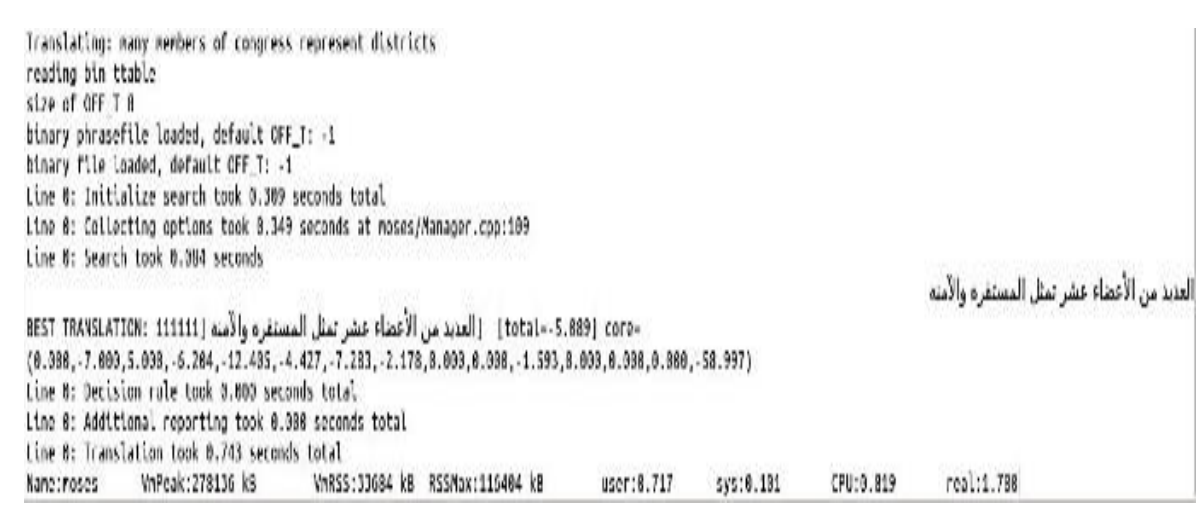

Fig. 4 Arabic translation of "Many members of congress represent districts" before tuning.

- The Arabic translation for the sentence "I will transfer the money" before tuning as shown in Fig. 5 is "الأول ستنقل إلى بغسل"

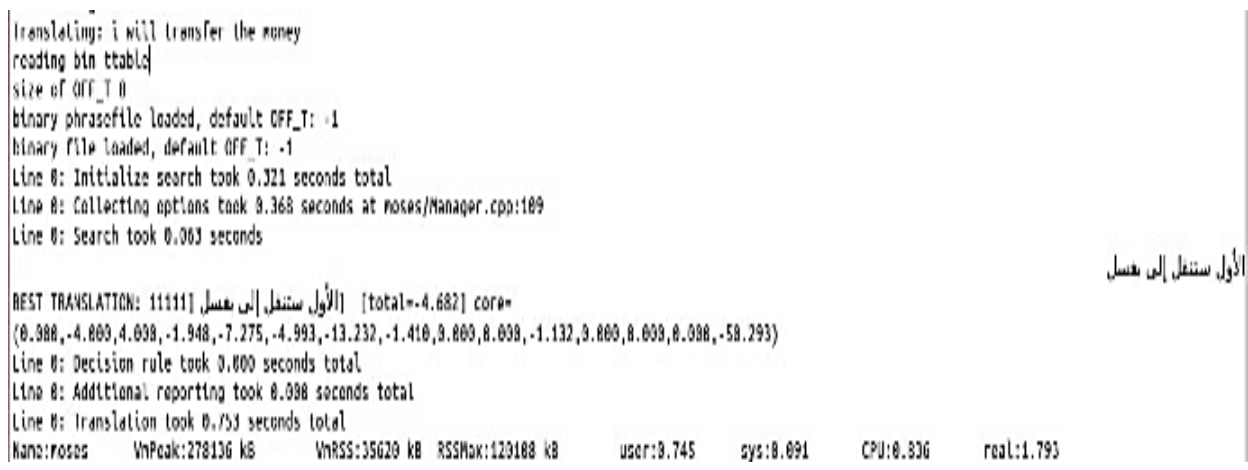

Fig. 5 Arabic translation of “I will transfer the money” before tuning.

- The Arabic translation for the sentence “Italy is far from Brazil” before tuning as shown in Fig. 6 is "إيطاليا عن البرازيل"

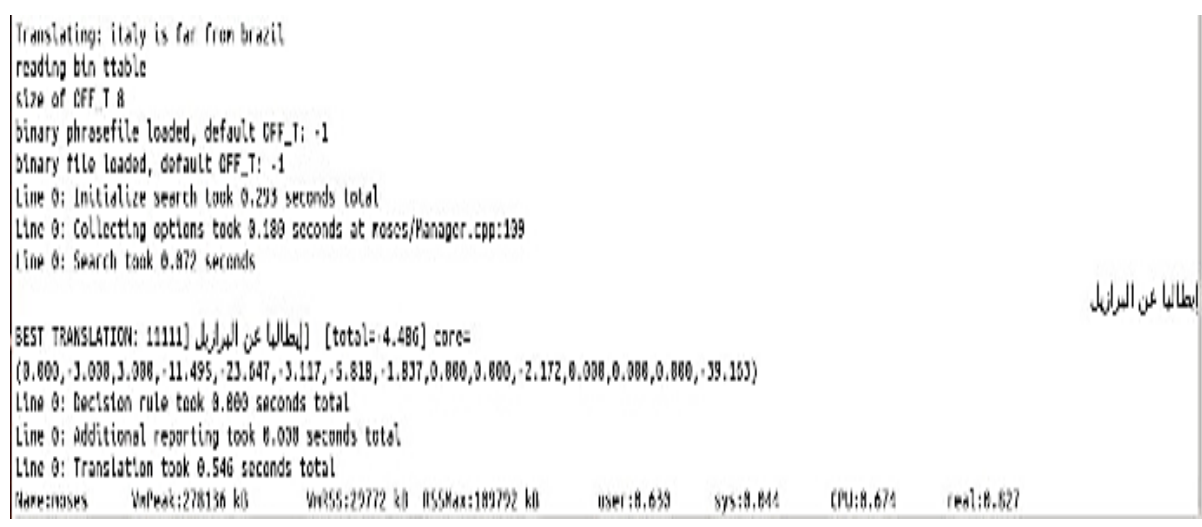

Fig. 6 Arabic translation of “Italy is far from Brazil” before tuning.

### 5.4.3 Model Translation after Tuning

- The Arabic translation for the sentence “Britain has largely excluded itself from any leadership role in Europe” after tuning as shown in Fig. 7 is "لقد استبعدت بريطانيا نفسها إلى حد كبير من الاضطلاع بأي دور قيادي في أوربا"

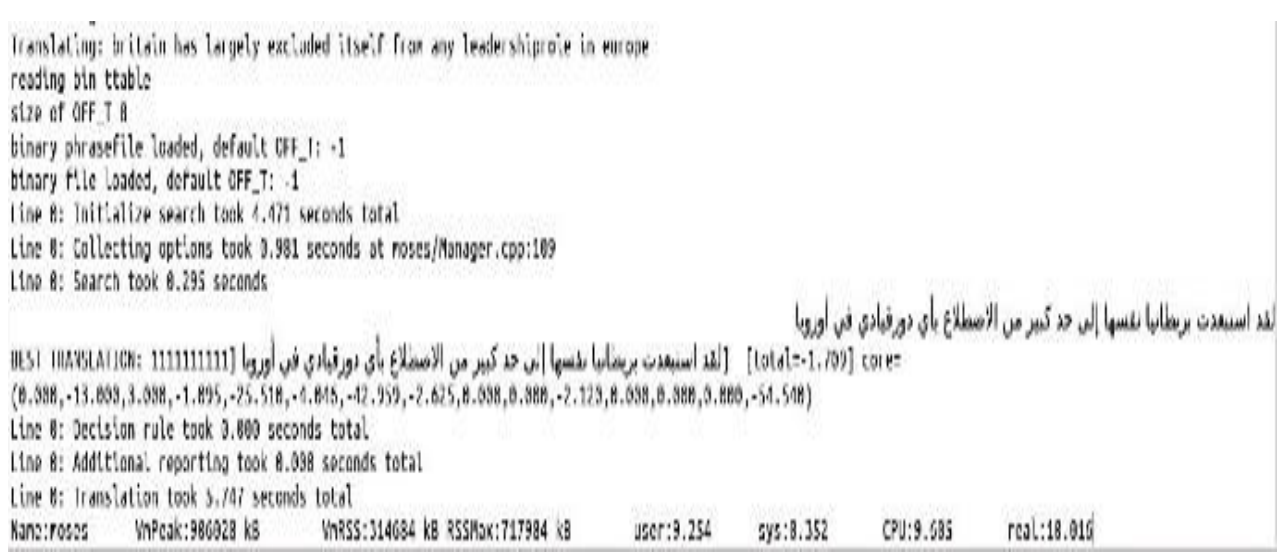

Fig. 7 Arabic translation of “Britain has largely excluded itself from any leadership role in Europe” after tuning.

- The Arabic translation for the sentence “Many members of congress represent districts” after tuning as shown in Fig. 8 is "إن العديد من أعضاء الكونجرس يمثلون المناطق"

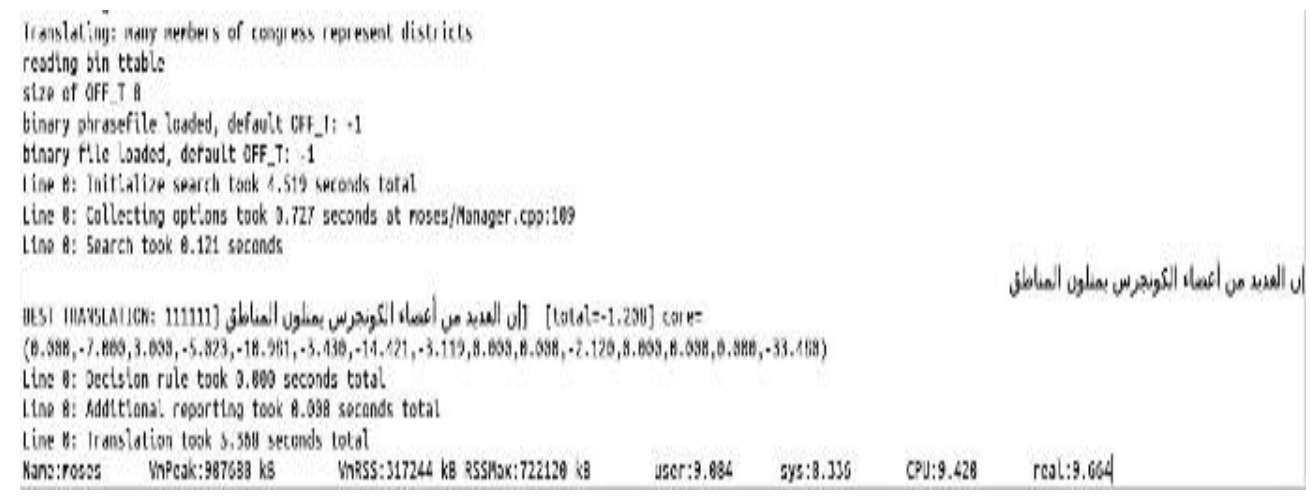

Fig. 8 Arabic translation of “Many members of congress represent districts” after tuning.

- The Arabic translation for the sentence “I will transfer the money” after tuning as shown in Fig. 9 is "سأحوّل المال"

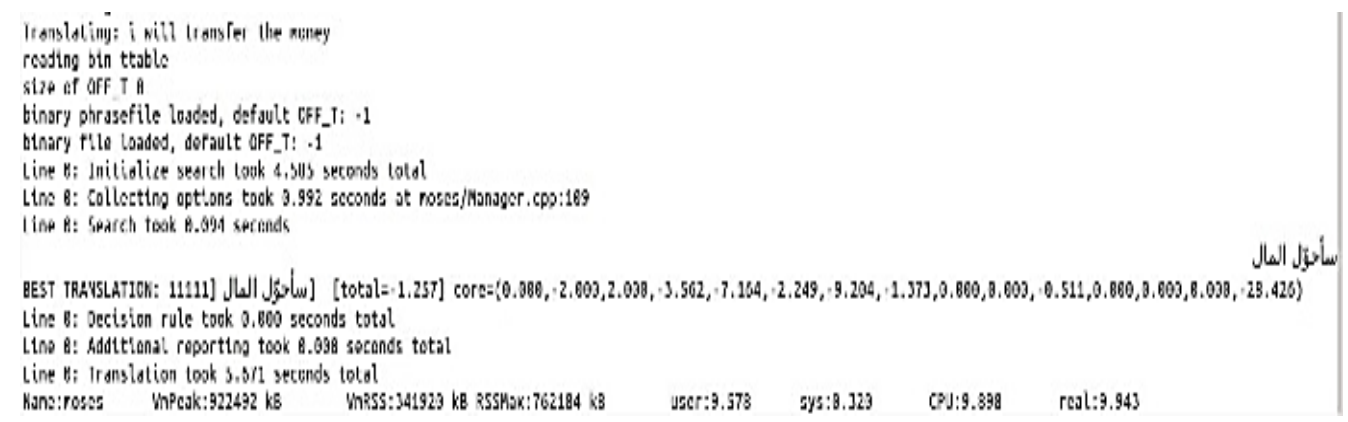

Fig. 9. Arabic translation of “I will transfer the money” after tuning.

- The Arabic translation for the sentence “Italy is far from Brazil” after tuning as shown in Fig. 10 is "إيطاليا بعيدة عن البرازيل"

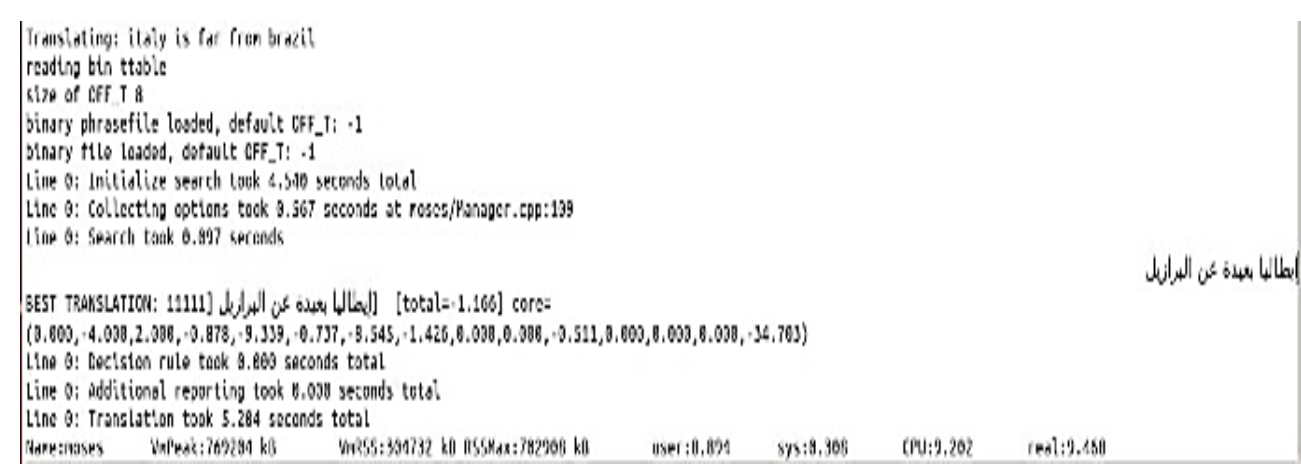

Fig. 10 Arabic translation of “Italy is far from Brazil” after tuning.

As shown from the previous translations, the translation before and after tuning are not the same. A huge improvement in translation noticed in translation after tuning than the translation before tuning. Tuning can be done for this model several times to improve the model quality, and many parallel corpuses can be trained to give better translation results. When comparing the model results with current SMT models such as Google it shows that there is a difference in the result between both systems, as the translation for the previous mentioned phrases in google translate are:

- The Arabic translation for the sentence “Britain has largely excluded itself from any leadership role in Europe” using Google

Translate is "وقد بريطانيا استبعدت نفسها إلى حد كبير من أي دور قيادي في أوروبا"

- The Arabic translation for the sentence "Many members of congress represent districts" using Google Translate is "العديد من أعضاء الكونجرس يمثلون دوائر انتخابية"
- The Arabic translation for the sentence "I will transfer the money" using Google Translate is "أنا سوف نقل الاموال"
- The Arabic translation for the sentence "Italy is far from Brazil" using Google Translate is "وايطاليا هي بعيدة كل البعد عن البرازيل"

After making comparison, it was found that Google generated translation for these sentences is less accurate from the proposed model generated translation. From these translation results, a conclusion can be noticed that the proposed model has enhanced some translation results rather than the current SMT models.

## 6. Conclusions

The proposed model in this paper incorporates efficiently many models at different levels namely: the language model, the alignment model, phrase based model, reordering model, translation model, and finally the tuning model. It was concluded that the achieved benefits of using these models are summarized as follows:

- The alignment models are used in statistical machine translation to determine translational correspondences between words in a sentence in one language with the words in a sentence with the same meaning in a different language.
- Language models increase the efficiency of the word alignment by using words depending on their context in the sentence.
- Phrase based model increases the capability of the proposed model by dealing with words and their correspondences or phrases in both languages at the same time. It adds value to the effectiveness of this model by translating a group of contiguous words in one language to a contiguous sequence of words in the other language.
- Reordering model uses all the mentioned previous models to generate reordering table by determining the orientation of two phrases based on word alignments at training time. This adds extra points to the reliability of this model and increases its dependability.

In addition, a set of other conclusions can be listed as follows:

- When combining different models in building and training statistical machine translation model, this will help in achieving good results even when using a small number of data.
- With model tuning, the translation results have been improved and the error rate has been minimized.
- The proposed model is adaptive, since new words, vocabularies, phrases, and sentences can be added and used in model training and tuning. This leads to the enhancement the Arabic translation quality.
- Context dependent language model achieves better and accurate translation. However, it is more costly.
- The order in which the used models are combined and trained has an effect in achieving better translation results.
- Generally, the implemented model proved by evaluation that it is a reliable, efficient, and effective in enhancing lexical statistical machine translation systems.

## 7. FUTURE WORK

The problem of machine translation has not solved yet. Much research and development still needed to earn the reliability of humans by preforming a fluency translation as humans do. The demand of faster and cheaper translation between languages will only increase with the need to share information between nations. Based on this work, expected future work and recommendations are as follows:

- Clustering is one of the most important techniques, which needs to be done on the Arabic language to solve the tokenization problem. Clustering as a general approach for dealing with issues of language sparsity and morphological analysis is promising.
- Use large amount of parallel corpus data to train the proposed model, to achieve better results.
- Use large amount of data for tuning from the same domain of the training data to get better tuning result.
- Every time this model is tuned, the better results it can achieve, so it would be better to tune it as much as possible.
- Use testing data from the same domain to get better actual results.
- Integration with other machine learning techniques such as artificial neural networks in model learning and training to achieve better results.

## References

[1] R. M. Zbib, "Using linguistic knowledge in statistical machine translation", Doctoral dissertation, Massachusetts Institute of Technology, 2010.

[2] R. Sarikaya and Deng, Y., "Joint morphological-lexical language modeling for machine translation", Proceedings of the Conference of the North American Chapter of the Association for Computational Linguistics; Companion

Volume, Short Papers, Association for Computational Linguistics, 2007, pp. 145-148.

[3] R. Zbib, S. Matsoukas, R. Schwartz, and J. Makhoul, "Decision trees for lexical smoothing in statistical machine translation", Proceedings of the Joint Fifth Workshop on Statistical Machine Translation and MetricsMATR, Association for Computational Linguistics, July 2010, pp. 428-437.

[4] A. R. Nabhan, and A. A. Rafea, "Tuning Statistical Machine Translation Parameters", Proceedings of IEEE International Conference on Information Reuse and Integration, 2005, pp. 338-343.

[5] P. Koehn, "Pharaoh: a beam search decoder for phrase-based statistical machine translation models", Proceedings of Machine translation: From real users to research, 6th Conference of the Association for Machine Translation in the Americas, Washington, DC, USA, Springer, Berlin Heidelberg, 2004, pp.115-124.

[6] K. Papineni, S. Roukos, T. Ward, and W. J. Zhu, "BLEU: a method for automatic evaluation of machine translation", Proceedings of the 40th annual meeting on association for computational linguistics, Philadelphia, July 2002, pp. 311-318.

[7] N. Habash, and O. Rambow, "Arabic tokenization, part-of-speech tagging and morphological disambiguation in one fell swoop", Proceedings of the 43rd Annual Meeting on Association for Computational Linguistics, June 2005, pp. 573-580.

[8] I. Badr, R. Zbib, and J. Glass, "Segmentation for English-to-Arabic statistical machine translation", Proceedings of the 46th Annual Meeting of the Association for Computational Linguistics on Human Language Technologies: Short Papers, Ohio, USA, June 2008, pp. 153-156.

[9] F. J. Och, and H. Ney, "Improved statistical alignment models", Proceedings of the 38th Annual Meeting on Association for Computational Linguistics, USA, October 2000, pp. 440-447.

[10] I. Badr, R. Zbib, and J. Glass, "Syntactic phrase reordering for English-to-Arabic statistical machine translation", Proceedings of the 12th Conference of the European Chapter of the Association for Computational Linguistics, Athens, Greece, March 2009, pp. 86-93.

[11] K. Heafield, P. Koehn, and A. Lavie, "Grouping Language Model Boundary Words to Speed K-Best Extraction from Hypergraphs", Proceedings of the Annual Conference of the North American Chapter of the Association for Computational Linguistics, Atlanta, USA, 2013, pp. 958-968.

[12] N. A. Smith, "Log-linear models. Department of Computer Science, Center for Language and Speech Processing", Johns Hopkins University, 2004.

[13] P. Koehn, "Statistical machine translation", Cambridge University Press, 2010.

[14] A. Rafalovitch, and R. Dale, "United Nations general assembly resolutions: A six-language parallel corpus", Proceedings of the MT Summit, Vol. 12, August 2009, pp. 292-299.

[15] J. Tiedemann, "Parallel Data, Tools and Interfaces in OPUS", In LREC, May 2012, pp. 2214-2218.

**Ahmed G. ElSayed Salama** Ahmed has a B.Sc. in Computer and Information Systems from Sadat Academy for Management Sciences of Cairo, Egypt (2007). Ahmed is a software quality control engineer with extensive experience and management skills and works for a hi-tech telecommunication company that develops and sells solutions that enable service providers to efficiently deliver high quality voice and data services over broadband access networks. As a senior software quality control engineer he is responsible for the quality software products produced by the company.

**Ahmed S. Salama** received his B.Sc. degree in computer sciences and information systems from Sadat Academy in 1991 Egypt, M.Sc. in Information Technology from Alexandria University in 1998, and his Ph.D. In information technology in 2004 from Alexandria University Egypt, He is mainly interested in machine learning applications, and knowledge discovery in Bid Data. He worked as an assistant professor in information systems department in faculty of computer sciences – Taiba University – KSA, and now is working as an assistant professor in computer and information systems department in Sadat Academy.

**Alaa El-Din M. El-Ghazali** received his Ph.D. in accounting from Suez Canal University Egypt, in 1993. He worked as a president of Sadat Academy for Management Sciences and now is working as a professor in computer and information systems department in Sadat Academy.